\documentclass[conference]{IEEEtran}
\IEEEoverridecommandlockouts
\usepackage{cite}
\usepackage{amsmath,amssymb,amsfonts}
\usepackage{algorithmic}
\usepackage{graphicx}
\usepackage{textcomp}
\usepackage{xcolor}
\usepackage{array}
\usepackage{lscape}
\usepackage{hyperref}
\usepackage[section]{placeins}
\def\BibTeX{{\rm B\kern-.05em{\sc i\kern-.025em b}\kern-.08em
T\kern-.1667em\lower.7ex\hbox{E}\kern-.125emX}}

\begin{document}
\pagestyle{plain}
\title{Making AI Evaluation Deployment-Relevant Through Context Specification}

\author{%
\IEEEauthorblockN{Matthew Holmes}
\IEEEauthorblockA{%
\textit{Intellect Frontier}\\%
UK \\
info@intellectfrontier.co.uk
}
\and
\IEEEauthorblockN{Thiago Lacerda}
\IEEEauthorblockA{%
\textit{Trustworks}\\%
USA \\ 
thiago@lacerda.com
}
\and
\IEEEauthorblockN{Reva Schwartz}
\IEEEauthorblockA{%
\textit{Civitaas Insights}\\%
USA\\%
\href{https://orcid.org/0000-0002-9012-6306}{https://orcid.org/0000-0002-9012-6306}%
}
\thanks{This work was conducted as part of the Forum for Real‑World AI Measurement and Evaluation (FRAME) at Virginia State University's Center for Responsible AI}
}

\maketitle

\begin{abstract}
With many organizations struggling to gain value from AI deployments \cite{MLQ2025StateOfAIinBusiness}, pressure to evaluate AI in an informed manner has intensified. Status quo AI evaluation approaches often mask the operational realities that ultimately determine deployment success,  making it difficult for organizational decision‑makers to know whether and how AI tools will deliver durable value. We introduce and describe context specification as a process to support and inform this decision making process. Context specification turns diffuse stakeholder perspectives about what matters in a given setting into clear, named constructs---explicit definitions of the properties, behaviors, and outcomes that evaluations aim to capture---so they can be observed and measured in context. The process serves as a foundational roadmap for evaluating what AI systems are likely to do in the deployment contexts that organizations actually manage.
\end{abstract}


\section{Introduction and Motivation: From Benchmarks to Decision-Grade Evaluation}

As organizationsal adoption of AI systems increases, decision makers are being asked to justify whether and how these technologies should be deployed and used in their operational settings. The most consequential impacts of AI unfold in real contexts and over time, reshaping organizational processes and incentives and producing a growing list of downstream societal impacts--both positive and negative \cite{SelgasCors2025SociotechnicalTransformation}. While many of these impacts arise from how people adapt to, misuse, or over‑rely on AI in specific settings, they are largely overlooked by status quo evaluation methods which remain oriented toward assessment and optimization of model capabilities\cite{bean_measuring_2025,wallach_position_2025,salaudeen_measurement_2025,weidinger2025eval}. As a result, organizational stakeholders lack evidence about what actually materializes in deployment environments and have little visibility into whether and how these technologies will generate genuine value or merely shift burdens to new parts of the workflow.\footnote{In this paper, the term "deployment" refers to the adoption and implementation of AI tools in users' settings.}

\subsection{The Need for Well-Defined Constructs}
Addressing the gap between current evaluation outcomes and what stakeholders need requires a shift from generic performance metrics toward explicitly defining the outcomes and behaviors that matter in deployment and translating them into concrete measurement targets--also referred to as "constructs". This process of turning diffuse ideas into precise, measurable constructs is known as construct systematization and is widely used in other measurement domains\cite{chouldechova_shared_2024}.

Techniques for construct development within AI evaluation currently fall into several categories. Some benchmarks overlook constructs entirely, treating scores as self-interpreting without specifying what real-world concept they are meant to capture \cite{bean_measuring_2025, salaudeen_measurement_2025}. Others import constructs from model tuning and optimization processes inside the AI stack, where labels like ``reasoning'' or ``helpfulness'' are treated as self-evident concepts rather than fully defined measurement goals \cite{chouldechova_shared_2024}. A complementary set of methods such as co-design, participatory techniques, and human-centered design are used to define locally grounded constructs related to values, harms, and preferences, but primarily to support system design rather thanevaluation \cite{world_economic_forum_ai_nodate,delgado_participatory_2023,mangold_design_2025,friedman_value_2002}.

A central question is whether the constructs used in AI design and development bear a stable relationship to what actually materializes in deployment. Even formalized constructs are often left empirically unexamined. It is unclear, for example, whether model‑level fairness metrics such as disparity measures can reliably track who benefits or is burdened in practice \cite{jacobs_measurement_2021,sap_social_2020,mulligan_this_2019,wallach_position_2025,citation-key}. Since many organizations adopting AI lack dedicated evaluation capabilities they must rely on readily available assessment tools, which are rarely designed for the questions they care about most \cite{yu2026aisafetybenchmarksbenchmark}. As a result, deployment and adoption decisions rest on numbers that look rigorous but may offer only weak guidance about particular deployments, including where they are likely to fail, or which communities will bear the risks.  

\subsection{Thesis and Problem Statement: Why Context Specification is Foundational}
Well-specified constructs offer a way out of this situation. Rather than inheriting metrics from model optimization or borrowing loosely from design methods, construct systematization provides a stable reference point for assessment. It can help organizations shift from trend-driven experimentation toward a clearly articulated destination and landmarks for whether, when, and how to deploy AI.

Context specification can be understood as a subtype of construct systematization that centers on “what matters” to those who adopt, use, govern, and oversee these technologies in deployment. This deployment-centered framing allows evaluators to direct their construct work toward how systems behave in the real world, rather than on model-centric abstractions\cite{Adcock_Collier_2001,chouldechova_shared_2024}.

We argue that systematic context specification is the foundational step for evaluating AI in the real world. It translates stakeholder priorities and contextual factors into evaluable constructs---explicit, named descriptions of properties, behaviors, and outcomes that can be observed and measured in a given setting. By linking these explicit constructs to candidate observables, context specification creates the conditions for gathering the systematic evidence that stakeholders need to manage and govern AI deployments.   It keeps evaluation in line with stakeholder goals and constraints, and focused on problems that actually materialize in practice rather than on abstract or hypothesized risks.

Without systematic context specification, evaluation tends to collapse into three recurring problems. First, effects are misattributed--observed outcomes can be interpreted as model performance when they may arise from a variety of contextual factors such as: workflow constraints, incentive structures, or human-system interaction patterns. Second, evaluation becomes reliant on brittle proxies that appear rigorous but drift from the real-world phenomena they are meant to capture. Third, deployment decisions end up based on metrics that lack a stable relationship to downstream impacts on people and institutions. Context specification addresses these problems not by prescribing particular metrics, but by clarifying what is being measured and why from the perspective of those in the deployment setting. 

\section{Conceptual Foundation: What Context Specification Yields}
The preceding section explains why context specification is necessary. This section clarifies what it produces as seen in Figure 1.

\begin{figure}[t]
  \centering
  \includegraphics[width=0.7\columnwidth]{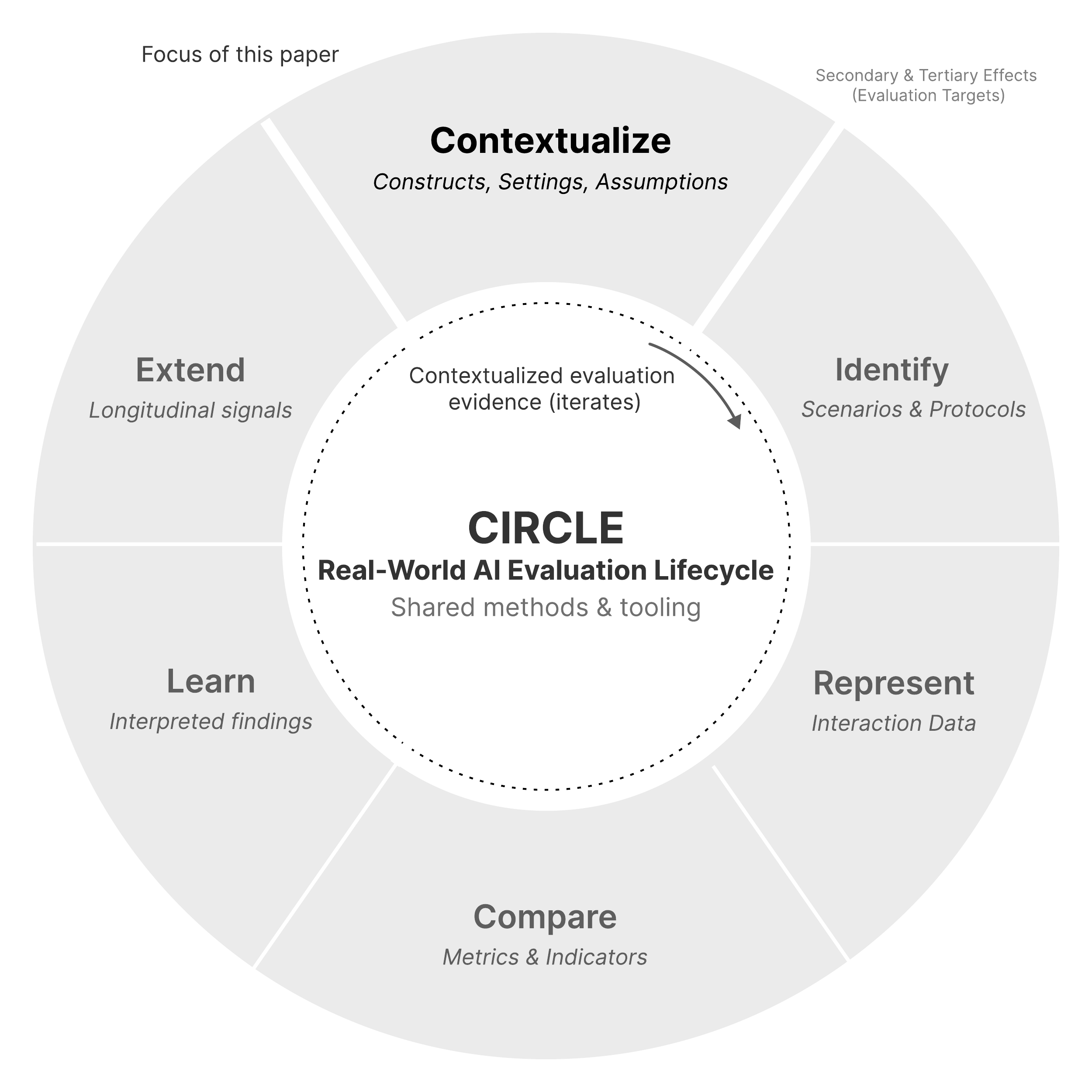}
  \caption{Context specification serves as the "Contextualize" step in the CIRCLE real-world AI evaluation lifecycle from\cite{realitycheck}.}
  \label{fig:circle-lifecycle}
\end{figure}

Systematic context specification uses stakeholder input to clarify what matters for an AI deployment in setting and then transform those initially diffuse concerns into structured measurement targets. It serves as a pre-step to selecting specific evaluation methods or metrics. The process yields a structured set of outputs that can be used to design an evaluation capable of supporting deployment-level claims. The outputs include:
\begin{itemize}
\item \textbf{Named stakeholder priorities}, articulated in relation to a specific deployment setting  and focused on the real-world impacts that matter to those who will live with the system’s consequences.
\item \textbf{Evaluable constructs}, defined as clear descriptions of properties, behaviors, or outcomes that can be meaningfully observed in that setting.
\item \textbf{Context-of-use elements}, including workflows, constraints, incentive structures, institutional norms, and likely use variants.
\item \textbf{Linking mechanisms}, which describe plausible pathways through which system behavior in context produces observable outcomes.
\item \textbf{Candidate observables and evidence needs}, distinguishing what can be inferred from model outputs (in silico) and what requires observation in deployment (in situ).
\item \textbf{Explicit assumptions and uncertainties}, identifying what is known, what is inferred, and what remains empirically open about which outcomes and behaviors matter in the deployment context.
\end{itemize}

A linking mechanism is a description of AI system behavior connected to its use by people and associated  real-world outcomes. For example, in a hiring workflow, an AI ranking model may influence human reviewers not only through its output scores but also by shaping which applications are surfaced first. The mechanism may involve cognitive shortcuts (e.g., defaulting to top-ranked candidates), workload pressures, or organizational incentives. Such pathways are not visible in model-level metrics but can be articulated during context specification to inform construct development and subsequent evaluation activities. The fact that AI systems shape human decision making is not inherently negative; the problem arises when these influences are poorly understood and unmodeled, leaving key aspects of AI’s real-world impacts unmeasured and unmanaged.

\section{Method: A Descriptive Process for Systematic Context Specification}
\label{sec:method_context_spec}
The method presented here is descriptive rather than prescriptive, and does not mandate particular controls or standards. 

Organizations can use context specification as a repeatable process for clarifying what is important in a deployment setting to ensure evaluation outcomes address their questions, such as: how a system could change existing workflows, which roles are affected, what risks arise, and how accountability is currently structured. The process is best informed by those who: 
\begin{itemize}
    \item make or shape system deployment and adoption decisions,
    \item will use, be accountable for or be affected by evaluation findings, and
    \item stand to be empowered or disadvantaged by how evaluation criteria are set\cite{Leslie,Bell,deshpande}. 
    \end{itemize}
    In line with theory of change approaches, the process is structured as Inputs $\rightarrow$ Activities $\rightarrow$ Outputs $\rightarrow$ Outcomes, making the assumed links between what the system does in context and the outcomes that matter to stakeholders explicit \cite{weiss1997theory,reuel2024generativeaineedsadaptive}. Teams conducting context specification activities should exhibit a blend of technical, measurement, participatory, and translational skills to effectively elicit local knowledge, anticipate evaluation requirements, and translate insights across the different disciplines in the AI evaluation ecosystem \cite{delgado_participatory_2023,Oduro2025TroublingTranslation,ParticipatoryTrust}

\subsection{Inputs}
Inputs ground the specification process in deployment-level realities by including detailed information about: who is participating in the process (stakeholders and roles, including decision makers, system users and operators, potentially affected individuals, and oversight functions), AI system purpose and anticipated contexts of use (including foreseeable variants), associated operational constraints and institutional norms, relevant documentation and logs, and known regulatory or governance touch points. 

\subsection{Activities}
Activities translate diffuse stakeholder inputs into a structured basis for measurement. They typically include (i) elicitation and synthesis to surface and refine stakeholder priorities and relevant contextual factors; (ii) systematization in which these are grouped, filtered, and articulated as candidate systematized constructs; and (iii) preliminary operationalization, in which plausible mechanisms and pathways are mapped to link system behaviors to observable indicators and downstream outcomes\cite{Adcock_Collier_2001,chouldechova_shared_2024,Kubisch1998ApplyingAT}.

\subsubsection{Elicitation modes}
Elicitation is used to collect diffuse input from relevant stakeholder communities for later organization and prioritization. This process can include in-person interviews and workshops, surveys, document review, and asynchronous exercises with relevant stakeholders that reflect organizational constraints and perspectives on the role of AI in their setting. Effective elicitation collects both explicit knowledge (documented policies, procedures, formal requirements) and tacit knowledge (experiential insights, uncodified workflow realities) to reveal "what matters" in a given setting and how that setting actually operates \cite{Popoola2024Advancements}

LLM‑driven approaches can be used to pre-structure the issue space before stakeholder engagement occurs, to stand in as a lightweight elicitation tool when direct engagement is constrained, or to iteratively prioritize and refine stakeholder feedback over time\cite{pasquale2025exploringusellmsrequirements}. Automated extraction and synthesis can summarize large volumes of documented knowledge (for example, surfacing system requirements and constraints, or edge cases from policies, logs, or job postings) and propose candidate topics for discussion\cite{leiva-araos_large_2025,aly2025evaluationlargelanguagemodels}. However, tacit knowledge---such as incentive structures, informal workarounds, and time pressure---typically requires direct stakeholder engagement. The method therefore treats automated extraction and human elicitation as complementary rather than interchangeable.

\subsection{Outputs}
The primary output of the process is the Context Brief, a structured artifact that clarifies and defines the priority items. It includes stakeholder details, intended and likely use contexts, constraints and norms, decision points (e.g., pilot go/no-go), candidate mechanisms and effects, key uncertainties, and evidence needs (including whether they are best addressed in silico or in situ, framed as questions of observability). The brief also provides an Item--Construct--Indicator mapping that translates stakeholder-articulated priority items about "what matters" into evaluable (or systematized) constructs. The context specification team maps the constructs to associated candidate indicators or prompts that can be tied to specific behaviors and outcomes for subsequent evaluation design and execution.

Context specification outputs do not constitute final metrics; rather, they establish the constructs and indicators that evaluations must address and why those targets matter, enabling clarity and transparency around what evaluation aims to capture. To support valid deployment‑relevant claims,  constructs must be collectively informed and described in ways that are technically precise, neutral, and unambiguous. This allows different actors to reliably interpret the meaning of the construct and associated evaluation outcomes in compatible ways.

\begin{figure}[t]
  \centering
  \includegraphics[width=\columnwidth]{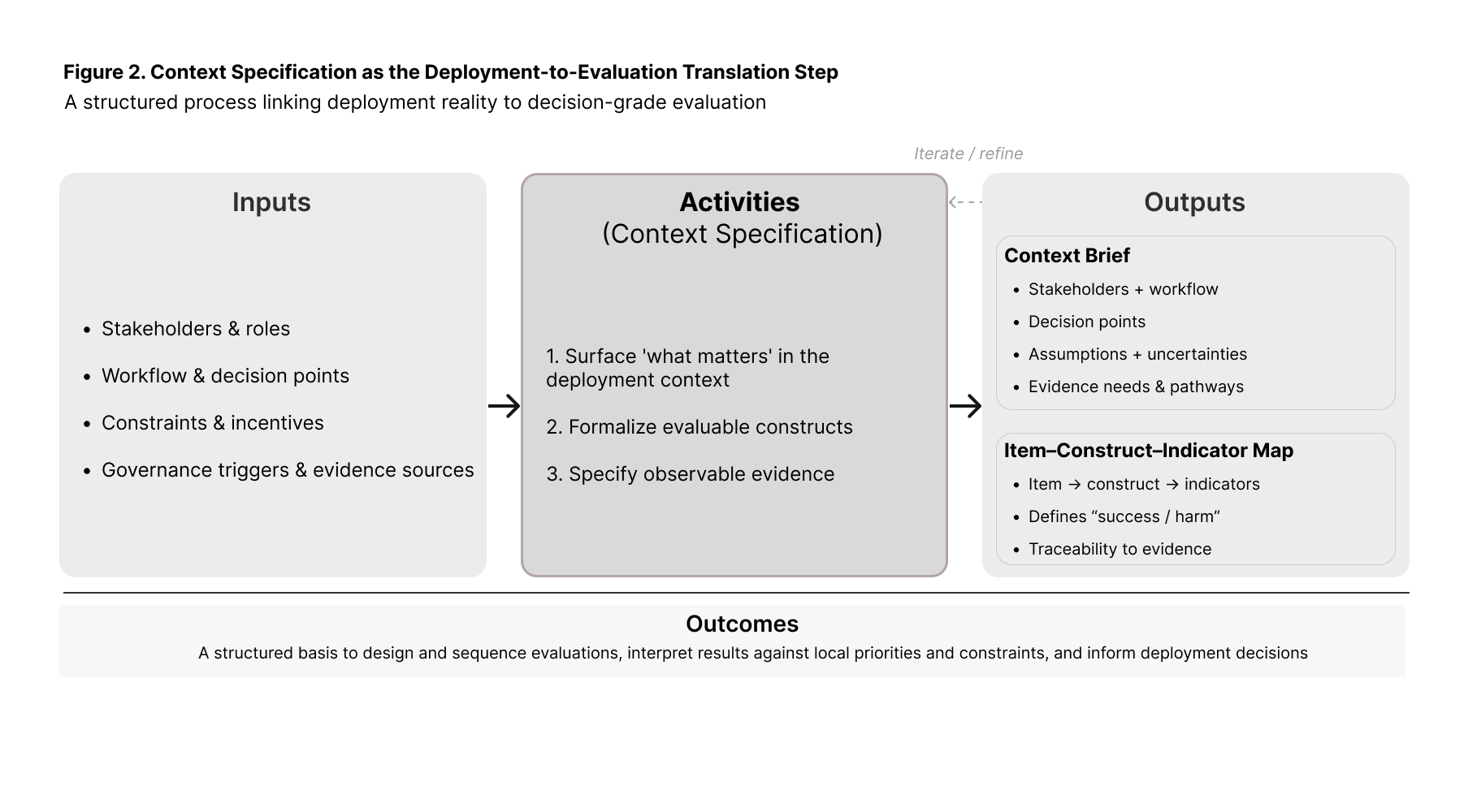}
  \caption{Context specification as the deployment-to-evaluation translation step: turning stakeholder priority items into evaluable constructs and evidence needs.}
  \label{fig:context-spec-translation}
\end{figure}

\subsection{Outcomes}
Context specification can help organizations shift how AI evaluation is understood and practiced for the longer term. First, it produces a clear, shared articulation of what matters to the relevant stakeholder community, including how success and harm are defined in a specific setting. Second, it generates a set of systematized constructs and associated indicators to structure evaluation design, rather than selecting methods on an ad hoc basis. Third, it creates an explicit record of key uncertainties and observability limits, clarifying which questions can be answered in silico, which require in situ observation, and where residual ambiguity may remain. Together, these outcomes equip stakeholders to design and sequence evaluations, interpret the resulting evidence in light of local priorities and constraints, and make downstream decisions such as AI pilot design, go/no-go determinations, scaling thresholds, or decommissioning. They also support institutional learning over time, as successive context briefs and evaluation cycles can be compared to track how items, constructs, and decision criteria evolve across deployments and sectors.

\subsection{Handoff to evaluation design choices}
The handoff to evaluation design occurs when the structured outputs constrain method choice. Constructs and linking mechanisms identified during context specification determine which phenomena must be observed in situ, which require longitudinal study, and which may be adequately approximated under controlled conditions. In this way, context specification informs---not replaces---subsequent evaluation planning.




\section{Example Use Case}

We leverage an example use case to illustrate how context specification can produce structured inputs that inform organizational governance and decision making about AI deployments.

\textbf{Use Case}:Publicly owned rail transport operator aiming to deploy an AI-driven HR screening system with chatbot functionality to support hiring decisions.

The AI-driven system combines predictive components (such as ranking candidates based on skills or generating a predicted job‑success score for each applicant) with a generative chatbot interface that HR professionals use to query and explore candidate information. This reflects a common real-world use case and demonstrates how AI behavior and its consequences are shaped by its use and the context in which it is deployed. In this case, task‑specific predictions feed directly into ranked lists and scores, and indirectly shape how HR staff talk and think about candidates during hiring discussions\cite{wang2026distinguishingtaskspecificgeneralpurposeai}. AI-based HR screening tools have already shown patterns of harmful bias that can influence downstream decisions and trigger operational and regulatory concerns\cite{Wilson2024GenderRA}. This example highlights how model behavior, organizational practices, contextual constraints, and user perceptions jointly determine real-world outcomes.  

\subsection{Methods}

We use Table~\ref{tab:concerns_constructs} and Table~\ref{tab:context_spec} to apply the methods outlined in Section~\ref{sec:method_context_spec} to the HR chatbot use case, demonstrating how the rail provider could use context specification to identify what they need to know about how the AI tool will behave in their environment, so they can make more informed decisions about whether and how to deploy it.

 Table~\ref{tab:concerns_constructs} displays how stakeholder-defined priorities from the use case can map to related constructs. 

\begin{table}[h]
\centering
\small
\begin{tabular}{|p{4cm}|p{3cm}|}
\hline
\textbf {Priority Items} & \textbf{Related Constructs} \\
\hline
Is this tool creating rework for my staff, or actually saving time? &
Productivity \\
\hline
Are employees over-relying on the tool and losing critical skills? &
Over-reliance \\
\hline
Is the tool shifting liability to our frontline workers? &
Accountability \\
\hline
Who is consistently benefiting from these systems, and who is absorbing new risks? &
Fairness and equity \\
\hline
\end{tabular}
\vspace{0.5\baselineskip}
\caption{Mapping Stakeholder-elicited Priorities to Related Constructs}
\label{tab:concerns_constructs}
\end{table}

Table~\ref{tab:context_spec} displays example inputs and outputs that may be generated during the context specification process. Inputs and activities are not fixed pairs, but can be reused and combined in ways that fit real‑world context and constraints.

Activities and inputs are selected using context‑sensitive elicitation choices: methods are chosen to fit the specific organizational setting, including existing workflows, constraints on time and resources, culture, and the particular properties of the AI system being evaluated, in line with community‑centered norm‑elicitation approaches\cite{Harris}.The activities for the use case are therefore selected in this spirit, reflecting the priorities surfaced by stakeholders rather than a rigid set of tasks.




\begin{landscape}
\begin{table}[p]
\centering
\small
\begin{tabular}{|p{2.5cm}|p{3cm}|p{4cm}|p{8cm}|}
\hline
\textbf{Input Category} & \textbf{Input Type} & \textbf{Activities (Description)} & \textbf{Example Output} \\
\hline
Organizational setting and constraints &
Org chart; basic HR metrics; public company information &
Process mapping of existing HR workflows and constraints for rail operations hiring, plus LLM‑assisted search to gather HR policies, recruitment documents, and prior decisions &
Mid‑sized rail operator with ~1{,}000 employees, hiring for Rail Operations Controller roles spread across ~20 regions; high‑volume, time‑pressured recruitment for safety‑critical posts \\
\hline
System, use case, and scope &
Vendor documentation for the screening model; procurement/business case; internal deployment note &
Wireframing to specify what the AI screening system and chatbot do, where they sit in the hiring flow, and how system outputs are leveraged and routed across the organization &
Tool Info: Third‑party HR screening tool that ranks applicants for Rail Operations Controller roles, marketed as improving speed and consistency in shortlisting; intended use limited to pre‑interview ranking and support for HR screeners\\
\hline
Evaluation purpose and scope &
Pre‑engagement materials such as email threads, briefings, and internal memos summarising why the review is being commissioned &
Interviews with HR staff, hiring managers, and senior leaders to clarify expectations, anticipated or already experienced system risks and benefits, and accountability structures &
HR staff had made note of how the AI‑enabled hiring process might alter risks in shortlisting; senior leadership decided to commission an evaluation to understand the risk and opportunity surface of the AI‑influenced Rail Operations Controller hiring process \\
\hline
Governance and compliance expectations &
Existing internal policies, procedures, and governance documents (both AI‑specific and general), plus any documented external regulatory or standards obligations &
LLM‑assisted search to identify applicable policies and guidance to link them to concrete constructs of interest such as fairness, over‑reliance, and productivity &
The company had identified several compliance obligations and governance standards for which understanding the deployment context and its potential impacts would be useful, motivating a closer look at how the AI‑enabled hiring process operates in practice \\
\hline
Stakeholders and roles &
Org chart; documents and emails related to the system procurement and how the system's use may impact the hiring process; informal knowledge about who is involved or affected &
Mock‑up of field testing for selected HR staff to see how scores and chat responses play out under realistic conditions &
A list of stakeholders and their roles is compiled, including decision‑makers, day‑to‑day users, and affected groups (e.g. HR professionals, senior leadership, and Rail Operations Controller candidates) \\
\hline
\end{tabular}
\vspace{0.4\baselineskip}
\caption{Context Specification for AI-Driven HR Screening: Inputs, Activities, and Example Outputs}
\label{tab:context_spec}
\end{table}
\end{landscape}

As the final artifact of the process, a brief sample of the Context Brief for the Rail Operations Controller hiring example follows:

\begin{itemize}
    \item \textbf{Stakeholders / roles:} HR screeners; hiring manager; operations representative responsible for coordinator performance.
    \item \textbf{Workflow / decision points:} System ranks candidates; HR may accept, override, or escalate rankings; chatbot used for borderline or surprising candidates and during hiring meetings.
    \item \textbf{Constraints / norms:} High application volume; time pressure; limited transparency into system logic; emerging norm of trusting top-ranked candidates by default.
    \item \textbf{Linking mechanisms:} Over-reliance on rankings under time pressure; chatbot summaries anchoring candidate framing; selection criteria potentially misaligned with disruption-management skills.
    \item \textbf{Uncertainties / evidence needs:} Frequency and direction of overrides; order of review (summary vs.\ application); whether overrides are recorded; gap between shortlisted skills and industry norms.
\end{itemize}

By following the inputs and activities summarized in Table~\ref{tab:context_spec}, context specification for this use case yields three main outcomes: a clear articulation of what matters to HR staff, hiring managers, and affected candidates; a set of systematised constructs (such as productivity, over‑reliance, shifting accountability, and fairness), and shared criteria for what counts as success or harm in AI‑enabled Rail Operations Controller hiring for the stakeholder organization. These outcomes provide a structured basis for selecting and combining evaluation methods and an explicit record of key uncertainties and observability limits. Together, these outcomes can support stakeholders in designing and sequencing evaluations, interpreting evidence about the screening tool, and making downstream decisions about piloting, go/no‑go, scaling, or decommissioning.

\section{Evaluation Design Choices as Tradeoffs}
Evaluation methods are not neutral instruments; they shape what can and cannot be observed. Once constructs and linking mechanisms are specified, method selection becomes a design decision involving tradeoffs between control and contextual richness.

Highly controlled methods (e.g., simulation, benchmarking, scenario modeling) offer clarity and reproducibility but often abstract away the contextual conditions through which downstream effects emerge. In contrast, high-context methods (e.g., field deployments, longitudinal observation, embedded studies) capture richer interaction effects but introduce variability that can require additional resources to constrain.

Context specification clarifies which constructs require which forms of evidence. For example, if a linking mechanism involves over-reliance driven by workflow pressures, purely in-silico methods will be insufficient and observational or field-based methods may be required. Conversely, constructs related to baseline system behavior independent of human interaction may be served by in silico testing methods alone.

Rather than prescribing a fixed hierarchy of methods, this framework positions method selection as contingent on the constructs identified during context specification. By making these dependencies explicit, organizations can design evaluation strategies that are proportionate to the risks and uncertainties present in their deployment context.


\section{Discussion: Limitations and Future Work}

\subsection{Example Use Case}
While designed to be realistic, the hypothetical use case provided in this paper cannot stand in for empirical observations or fully capture the organizational dynamics, politics, or historical dependencies that shape real deployments. It also focuses on a single high‑risk public‑sector context where decision making may lean more on the side of governance and compliance as opposed to operational innovation.

Future work should apply the approach in real deployments, to see how activity selection, elicitation, and context‑brief production work under actual time, staffing, and budget constraints. It should span multiple domains and risk profiles to test how far the approach transfers, which parts generalize, and where adaptation is needed for different institutional and sectoral settings.

Related work is currently being developed through the Forum for Real-World AI Measurement and Evaluation (FRAME) at Virginia State University's Center for Responsible AI. FRAME focuses on generating systematic evidence about AI-in-use across real-world settings through large-scale testing and structured observation. Its aim is to develop the infrastructure, methods, and evidence base needed to evaluate AI deployments in realistic settings, in support of organizational decision-making.

\subsection{Execution capability and culture}
Real-World examples of AI impact assessments suggest that organizations can lack the in‑house expertise needed to design and run robust, context‑aware elicitation processes\cite{ECNL_DIHR_FRIA_Guide_2025}, weakening the quality of such assessments. 

The effectiveness of context specification requires clear elicitation protocols \cite{Wahbeh}, the inclusion of stakeholders with appropriate local knowledge, and a broad set of socio-technical skills across the team conducting the exercise. In many organizations, developing and applying this kind of capability remains a challenge.

Further research should characterize the capability gap for conducting context specification across organizations and sectors, including typical needs for upskilling, role design, and external support. Empirical work should also examine how organizational politics, incentives, and culture limit what can be elicited, agreed, and recorded, and how this in turn affects the quality and defensibility of context briefs.

\subsection{Data and documentation preconditions}

The method presupposes that internal artifacts (for example, policies, process maps, and incident logs) exist and are of sufficient quality to support context specification. Reviews of AI documentation and risk-governance practice find that documentation is frequently incomplete, inaccurate, or not up to date due to incentive, resourcing, and workflow constraints, limiting how far contextual analysis and risk management can go in practice\cite{winecoff2024improvinggovernanceoutcomesai}.

Future work should examine how variability in internal data and documentation quality affects feasibility and outputs, including how organizations can scope, phase, or compensate when documentation is incomplete, inconsistent, or unreliable.

\subsection{Construct maturity and measurement readiness}

Context specification defines constructs of interest (such as over‑reliance and sycophancy) to ease translation of contextual priorities into evaluable targets. As these techniques are still nascent practice in the AI evaluation ecosystem, construct sets and associated measurement targets remain incomplete and may omit relevant risk or opportunity dimensions\cite{chouldechova_shared_2024}. Initiatives such as FRAME are intended to support the future development of shared construct libraries and scoring rubrics for AI-in-use.

Finally, further work should refine and expand the construct set and associated measurement targets captured in context specification activities, and assess their validity across domains and deployment conditions. This may include developing taxonomies, thresholds, and instrumentation that surface socio‑technical risk and opportunity more systematically over time.

\section{Conclusion}
The current AI evaluation ecosystem reflects its roots in system design and development. While useful for optimizing and comparing model capabilities, it provides limited insight for stakeholders deciding whether and how to deploy these technologies in their own settings. This paper introduces a context specification process for identifying "what matters" in deployment environments so that key constructs can be developed to guide system evaluation and inform deployment decisions at the organizational level. By connecting evaluation criteria to stakeholder-defined constructs, context specification creates a new pathway for decision-makers to obtain the information they need, facilitating responsible AI practice and enhancing the value of AI technologies in operational contexts.

\section{Definitions of Relevant Terms}
\textbf{Construct}: An abstract, latent concept or theoretical variable, not directly observable, that is defined for scientific purposes and measured indirectly through multiple observable indicators or items.

\textbf{Construct operationalization}: Linking a systematized concept to appropriate indicators and scores in a coherent measurement framework \cite{Adcock_Collier_2001}.

\textbf{Construct systematization}: The process of clarifying and organizing a concept by specifying its meaning, dimensions, and relationships to other concepts\cite{Adcock_Collier_2001}.

\textbf{Evaluation}: (1) Systematic determination of the extent to which an entity meets its specified criteria; (2) Action that assesses the value of something\cite{ISOIECIEEE24765_2017}.

\textbf{In silico testing}: Testing or experimentation carried out entirely on a computer, using computational models and simulations.

\section{AI Use Disclosure}
This manuscript was prepared with targeted assistance from AI technology. AI was used to (i) generate and format some BibTeX citation entries, (ii) draft and refine LaTeX table code and cross-references, and (iii) provide limited support for restructuring and line-editing sections of the text (e.g., smoothing phrasing, tightening paragraphs). All conceptual contributions, methodological design, analysis, arguments, and final editorial decisions were made by the authors, who also reviewed and edited all AI-assisted content for accuracy and appropriateness.

 \bibliographystyle{IEEEtran}
\bibliography{references}

\

\end{document}